%% file: main.tex
\documentclass[sigconf]{acmart}
\AtBeginDocument{%
  \providecommand\BibTeX{{%
    \normalfont B\kern-0.5em{\scshape i\kern-0.25em b}\kern-0.8em\TeX}}}

\copyrightyear{2024}
\acmYear{2024}
\setcopyright{rightsretained}
\acmConference[Websci '24]{ACM Web Science Conference}{May 21--24, 2024}{Stuttgart, Germany}
\acmBooktitle{ACM Web Science Conference (Websci '24), May 21--24, 2024, Stuttgart, Germany}\acmDOI{10.1145/3614419.3644001}
\acmISBN{979-8-4007-0334-8/24/05}

\usepackage{xcolor}

\usepackage{url}
\usepackage{csquotes}
\usepackage{amsthm}
\usepackage{tikz}

\newtheorem{definition}{Definition}

\usepackage{booktabs}
\usepackage{multirow}
\newcommand{\llmUsed}{Llama-2-70b-chat-hf}
\let\emptyset\varnothing
\begin{document}
\title[Lost in Recursion: Mining Rich Event Semantics in Knowledge Graphs]{Lost in Recursion: Mining Rich Event Semantics \\in Knowledge Graphs}
%
\pdfstringdefDisableCommands{%
  \def\\{}%
}
\author{Florian Plötzky}
\email{ploetzky@ifis.cs.tu-bs.de}
\orcid{0000-0002-4112-3192}
\affiliation{%
  \institution{Institute for Information Systems\\ TU Braunschweig}
  \streetaddress{Mühlenpfordtstr. 23}
  \city{Braunschweig}
  \country{Germany}
  \postcode{38106}
}  

\author{Niklas Kiehne}
\email{kiehne@ifis.cs.tu-bs.de}
\orcid{0009-0009-1564-9340}
\affiliation{%
  \institution{Institute for Information Systems\\ TU Braunschweig}
  \streetaddress{Mühlenpfordtstr. 23}
  \city{Braunschweig}
  \country{Germany}
  \postcode{38106}
}   

\author{Wolf-Tilo Balke}
\email{balke@ifis.cs.tu-bs.de}
\orcid{0000-0002-5443-1215}
\affiliation{%
  \institution{Institute for Information Systems\\ TU Braunschweig}
  \streetaddress{Mühlenpfordtstr. 23}
  \city{Braunschweig}
  \country{Germany}
  \postcode{38106}
}   
\renewcommand{\shortauthors}{Florian Plötzky, Niklas Kiehne, and Wolf-Tilo Balke}
\begin{abstract}
Our world is shaped by events of various complexity.
This includes both small-scale local events like local farmer markets and large complex events like  political and military conflicts.
The latter are typically not observed directly but through the lenses of intermediaries like newspapers or social media.
In other words, we do not witness the unfolding of such events directly but are confronted with \emph{narratives} surrounding them.
Such narratives capture different aspects of a complex event and may also differ with respect to the \emph{narrator}. 
Thus, they provide a rich semantics concerning real-world events.
In this paper, we show how narratives concerning complex events can be constructed and utilized.
We provide a formal representation of narratives based on \emph{recursive nodes} to represent multiple levels of detail and discuss how narratives can be bound to event-centric knowledge graphs.
Additionally, we provide an algorithm based on incremental prompting techniques that mines such narratives from texts to account for different perspectives on complex events.
Finally, we show the effectiveness and future research directions in a proof of concept.
\end{abstract}

\maketitle
\section{Introduction}
\label{sec:introduction}
\input{sections/introduction}

\section{Events and Narratives}
\label{sec:formalization}
\input{sections/narratives}

\section{Mining and Binding Narratives}
\label{sec:event_structures}
\input{sections/event_structures}

\section{Proof of Concept} 
\label{sec:proof_of_concept}
\input{sections/proof_of_concept}

\section{Related Work}
\label{sec:related_work}
\input{sections/related_work}

\section{Conclusion}
\label{sec:conclusion}
\input{sections/conclusion}

\begin{acks}
Supported by the Leibniz-ScienceCampus Postdigital
Participation funded by the Leibniz Association (Leibniz-Gemeinschaft).
\end{acks}
\bibliographystyle{ACM-Reference-Format}
\bibliography{biblio}
\end{document}

%% file: sections/introduction.tex
With the rise of the Web, research concerning large and complex events has become necessary to connect the dots and gain insights into the intricacies of such events \cite{leban2014eventregistry}.
However, most events that occur are observed through the lenses of Web content like online news and social media.
As an effect, we mostly do not observe the event itself but are presented with  \emph{narratives} about it.
Consequently, journalists, analysts, and political scientists are often confronted with different and potentially incompatible stories regarding a specific event, which must be compared and unified \cite{baden2018viewpoint,frermann2023narrativeframes}.
Such tasks are aggravated by the usage of framing techniques \cite{frermann2023narrativeframes,opdahl2021ontologiesjournalisticangles} and patterns of strategic communication \cite{wilson2018strategicnarratives}.
In either case, systems enabling the representation and comparison of narratives regarding complex events are beneficial to uncover different perspectives on such events and unpack the surrounding narratives.

\begin{figure*}
    \includegraphics[width=.85\linewidth]{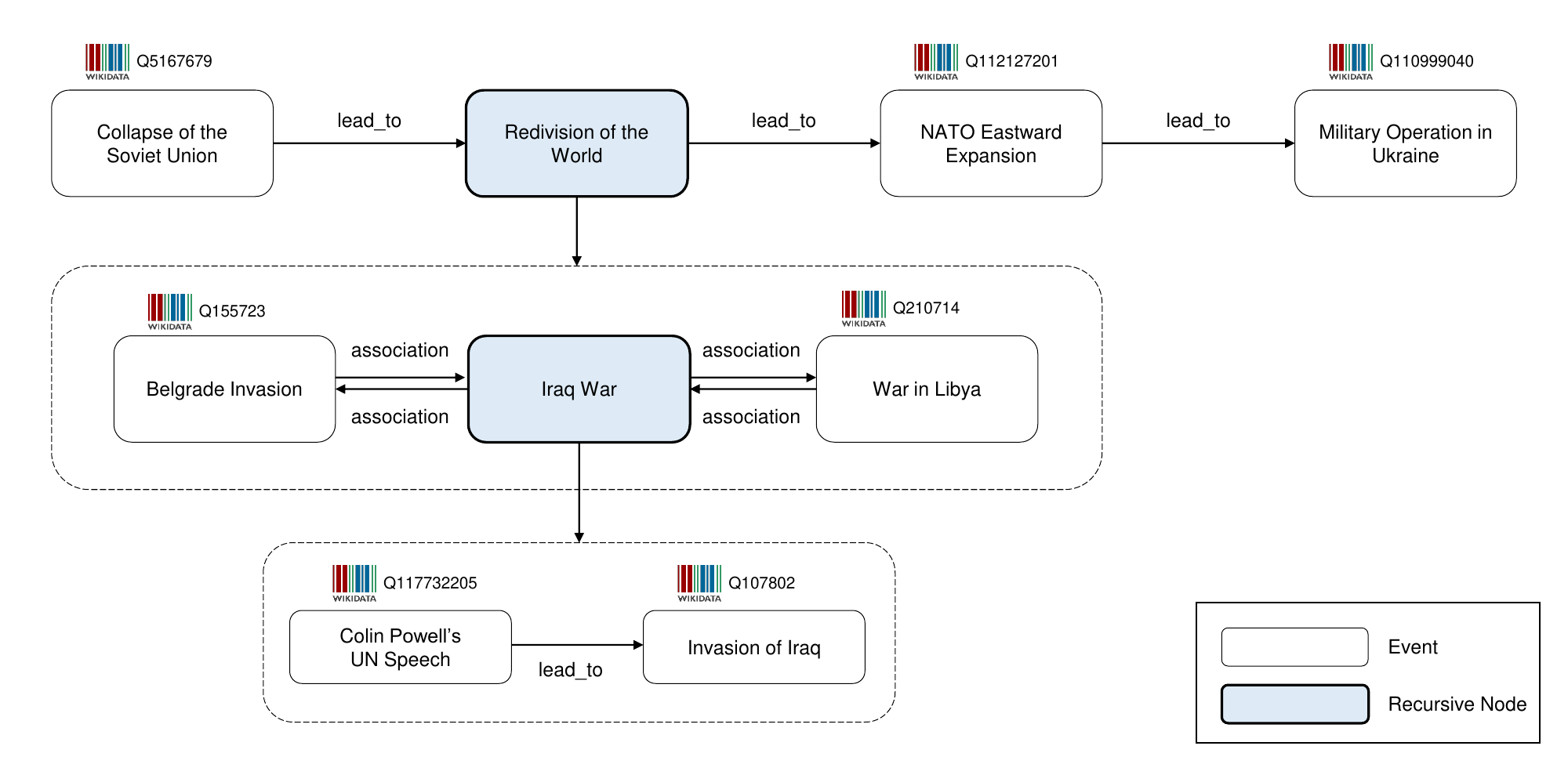}
    \caption{A simplified narrative that depicts Vladimir Putin's address to the nation. The events \enquote{Redivision of the World} and \enquote{Iraq War} are \emph{recursive nodes} in this example and reflect the inner structure of those events as exemplified in Putin's speech. All events also available in Wikidata are marked with their respective identifier.}
    \label{fig:ukraine_invasion_example}
\end{figure*}

At the core, we understand a narrative as a sequence of events arranged by a \emph{narrator} to describe properties and inner workings of complex events.
For instance, an event like a war usually comprises different battles, an event that started the war (which is not necessarily a battle), and an event that ended it.
Additionally, narratives can be used to describe properties and justify specific events.
Take, for instance, the narrative told by Vladimir Putin in his address to the nation at the advent of the Ukraine invasion in 2022.\footnote{\url{https://www.bloomberg.com/news/articles/2022-02-24/full-transcript-vladimir-putin-s-televised-address-to-russia-on-ukraine-feb-24}, last accessed Feb. 15, 2024}
Here, Putin connects over 30 years of events in a specific way, introduces new events like the redivision of the world (i.e., different wars involving Western countries since the Collapse of the Soviet Union), and frames the role of participants like NATO in a specific way. Fig.~\ref{fig:ukraine_invasion_example} shows an excerpt of this narrative in a structured way that will also be our running example throughout the paper.

Ultimately, this whole narrative is crafted to justify the invasion.
Obviously, parts of it are framed and adjusted to achieve this objective (cf., \cite{fisher2022putinscaseforwar} for a critical discussion and a different point of view).
Hence, a narrative constructed from a more Western perspective like that of the United States would be different at large, especially regarding the motivation of the NATO eastward expansion in that narrative.
Still, both narratives would concern the same named event, i.e., the  Russo-Ukrainian War, regarding the war's cause.
The differences and similarities of such narratives, however, may expose specific highlighting of events and, therefore, hint at the intentions of the respective narrator.

In this paper, we focus on how narratives of complex events can be used to describe the inner structure of an event from different points of view (i.e., by different narrators).
Specifically, we refer to the idea of the faculty of language \cite{hauser2002faculty} to construct narrative events chains of different granularity.
The core idea is that since human language is constructed recursively, narratives can also be seen as the result of compositions of different narratives, where a narrative is simply the composition of at least two events \cite{kroll2020narratives}.
Putin's narrative is a composition of narratives concerning the collapse of the Soviet Union, the events around the NATO eastward expansion, and the redivision of the world as depicted in Fig.~\ref{fig:ukraine_invasion_example}.
All of those different pieces must, of course, fit (from the narrator's view) to create a unified narrative of the particular event in question, i.e., the Russo-Ukrainian War.

Of course, representing complex events is not a new topic.
Most research in this direction was conducted in works concerning ontological aspects of events \cite{scherp2009eventmodelf,guizzardi2013towardsontologicalevents,almeida2019eventsasentities} and put in practice by the implementation of event-centric knowledge graphs (ECKGs) like EventKG~\cite{gottschalk2018eventkg} and OEKG~\cite{gottschalk2021openeventkg}.
While some recent works introduced narrative aspects \cite{ploetzky2022narrativeaspects,porzel2022narrativizingkgs} and viewpoint-specific annotations \cite{ploetzky2023attributions}, narratives are not only characterized by role-assignments to participants \cite{frermann2023narrativeframes} but also by the choice and arrangement the events discussed in it.
Putin's speech, for instance, arranges a sequence of events as an explanatory factor for the invasion of Ukraine (from his perspective).
However, a corresponding ECKG may not directly reflect the narrative.
That is, a narrative may combine otherwise unrelated events, for instance in a causal relationship, with respect to a specific \emph{viewpoint}.
Therefore, such relationships are only valid from a specific point of view but not generally agreed upon, and hence, they are not materialized in ECKGs.

In addition, narratives can be organized hierarchically using \emph{recursive nodes}, e.g., \enquote{Redivision of the World} involves three distinct and usually unconnected events.
Another application for such recursive nodes is to capture a specific aspect a narrator wants to highlight.
For our running example, this is true for the Iraq War, where the (in hindsight false) allegations of the Iraqi possession of weapons of mass destruction (WMD) in a meeting of the UN Security Council are explicitly highlighted~\cite{borger2021colinpowell}.
A narrative can be seen as an \emph{overlay} for ECKGs that connects and constructs subgraphs from the KG and enriches it with additional (potentially missing) events only available in the narrative.
By grounding narratives in well-known ECKGs, we can express and compare different perspectives on complex events and reason about, for instance, similarities, differences, and variants of different narratives.
Hence, we argue that narratives add a rich semantic to ECKGs.

In this paper, we adopt and extend the narrative model of Kroll et al.~\cite{kroll2020narratives} specifically to represent narratives concerning real-world events in a recursive manner.
We contribute a method to mine narratives from different perspectives and on multiple levels based on state-of-the-art event schema induction techniques~\cite{li2023schemainductionprompting}.
Additionally, we discuss how such mined events should be linked to ECKGs.
Finally, we demonstrate the effectiveness of our approach in a proof of concept.

%% file: sections/narratives.tex
In this section, we first introduce events, ECKGs, and viewpoints to represent narrative aspects of ECKGs.
We then provide a formalization of narratives, which is used throughout this paper.

\subsection{Events}
\label{subsec:events}
In general, events connect entities (the participants of the event) and set them into a temporal and spatial context.
We define events in a formal way as follows:

\begin{definition}[Events]
    \label{def:events}
    Given a set of entities $\mathcal{E}$, a set of locations $\mathcal{L}$ (which could either be entities or literal values), and literals referring to points in time $\mathcal{T}$, 
    an \emph{event} is an interaction between participants $p \in \mathcal{E}$ that takes place at a given location $l \in \mathcal{L}$ and at a specific point in time $t \in \mathcal{T}$ or in a specific time span $(t_1, t_2) \in \mathcal{T}\times\mathcal{T}$.
    The set of events is denoted by $\Gamma$.
\end{definition}

Optionally, events can be described by an \emph{event type} that describes the typical shape of an event, e.g., what kind of participants are involved.

Of course, this definition is quite broad.
That is, it can fit a small-scale event like a local farmer's market with a few hours duration and a year-long war like the ongoing war between Russia and Ukraine.
The latter events are often drilled down to multiple, simpler ones.
For the Russo-Ukrainian War we can, for instance, consider the Snake Island Campaign~(Q111012552) or the Battle of Chernihiv~(Q111013915), both in 2022, as parts of this event.
To account for such complex events, we define them accordingly.

\begin{definition}[Complex Events]
    \label{def:complex_events}
    A \emph{complex event} is an event that is connected to any other event by either a mereological relationship (i.e., a \emph{part-of} relation) or a hierarchical relationship (i.e., a \emph{sub-event} relation).
\end{definition}

Current ECKGs primarily reflect on complex events by sub-event relationships as first introduced on a larger scale in EventKG~\cite{gottschalk2018eventkg}.
For each event that is part of a complex event, the participants, location, and time must fit into the analogous components of the complex event.
That is if the complex event has a time span of one year, the time of each sub-event must be located in this span.
With those definitions in place we can formally introduce event-centric knowledge graphs.

\begin{definition}[Event-Centric Knowledge Graphs]
   An \emph{Event-Centric Knowledge Graph} is a knowledge graph (i.e., a triplestore consisting of subject-predicate-object triples) where the majority of triples contain a (complex) event in their subject or object respectively.
\end{definition}

Typically, such knowledge graphs are constructed by using the well-known Resource Description Framework \cite{klyne2014rdfprimer}.
Besides knowledge graphs explicitly designed as ECKSs (e.g., EventKG and OEKG), this definition also encompasses portions of general-purpose knowledge graphs like Wikidata~\cite{vrandecic2014wikidata}.
The latter case was already illustrated in~\cite{rudnik2019newseventkg}.
Events in such graphs can be seen as a special form of an entity in the graph that adheres to Def.~\ref{def:events}, i.e., that connects other entities and bounds this interaction in space and time.
Entities are included in such graphs mostly as participants for events.
Additionally, they can be further characterized by additional attributes that are not bound to their participation in an event (e.g., typical attributes like names, birth dates, and nationalities).

\subsection{Viewpoints}
\label{subsec:viewpoints}
ECKGs offer structured event representation at scale.
In contrast to textual event repositories (e.g., GDELT~\cite{leetaru2013gdelt} and Event Registry~\cite{leban2014eventregistry}), they also offer a canonical representation of named events and allow for querying and reasoning about events.
However, transforming event knowledge from unstructured sources into ECKGs tends to miss \emph{narrative aspects} for events, e.g., specific roles like an \emph{aggressor} in a conflict.
Such aspects can be problematic since it might be disputed whether an event participant can be seen as, for instance, an aggressor.
Tackling this problem, we recently introduced \emph{viewpoints}, i.e., the representation of disputed facts from multiple perspectives to ECKGs \cite{ploetzky2023attributions}.
We rely on this work and define viewpoints as follows.

\begin{definition}[Viewpoints]
   A \emph{viewpoint} represents the consensual stance (valid or invalid) towards a specific event-centric claim that is expressed by a group of entities.
   Each entity in such a group of entities $G$ must be capable of expressing a stance.
   We denote the set of viewpoints as $\mathcal{V}$. 
\end{definition}

At the core, specific claims concerning an event (e.g., the role a participant played in the event) might be disputed, in terms of their validity, by different groups of entities $G$. 
A viewpoint aggregates the individual stances of $G$ regarding the validity of this claim into a consensual stance for this group, e.g., by a majority vote.
Examples for typical group members include politicians, newspaper columnists, and speakers of governmental and non-governmental organizations.

Viewpoints belong to a specific viewpoint model that determines whether two viewpoints $v_1, v_2 \in \mathcal{V}$ can be subsumed, i.e., whether a claim that is valid in $v_1$ can also be seen as valid in $v_2$.
A simple variant of such a viewpoint model is a group hierarchy that determines which groups $G_1, G_2$ can be aggregated to form a larger group $G^*$.
For instance, the stances of the President of the United States and the US Congress may be aggregated into the \enquote{US viewpoint}.
If two viewpoints $v_1$ and $v_2$ can be aggregated according to such a model, we call them \emph{viewpoint-compatible}.
Additionally, we call predicates in ECKGs that are annotated with viewpoints \emph{attributions}.
An example is depicted in Fig.~\ref{fig:param_pred_eckg} between Event 3 and an entity.

\subsection{Narratives}
\label{subsec:narratives}

In this section, we finally introduce narratives as a representation to connect events, for instance, in ECKGs meaningfully.
As a motivation, we refer to the recent idea of hierarchical event schemas as presented in \cite{li2023schemainductionprompting}.
Basically, complex event types like wars or pandemic outbreaks are described by hierarchical schema graphs.
A schema graph connects the most important entities and sub-events of an event type to give an abstract description of what typically happens in those scenarios.
This can be seen as a hierarchical schema since each sub-event also spans an event schema graph that can be traversed recursively downwards.

We argue that recursion in such schemas is a valuable mechanism for the deep representation of otherwise complex events.
Especially in conflict situations, understanding the root causes of how the actual conflict played out can help researchers understand the actual mechanism behind the conflict for purposes of categorization and comparison with other events.

In comparison, narratives enhance this idea in two aspects.
First, while event schemas aim to describe the schema of a typical representative of an event, real-world events are often convoluted or surrogate events.
Take, for instance, the \enquote{Redivision of the World} in Fig.~\ref{fig:ukraine_invasion_example}.
This event comprises a number of real-world events but is constructed in the narrative itself.
Hence, there will be no counterpart in an ECKG besides the sub-events like the Belgrade invasion, which can be found in ECKGs and other event-centric repositories.
In addition, convoluted events are those that do not follow a typical schema.
This can be seen by the NATO eastward expansion that comprises multiple steps and can hardly be categorized by a general schema.
Second and in addition to Sect.~\ref{subsec:viewpoints}, especially large-scale events like wars, the choice and connotation of events depends on the narrator up to a certain degree.
That is, depending on the perspective chosen, specific events are highlighted or framed in a way.
In Fig.~\ref{fig:ukraine_invasion_example}, this is visible for the Iraq War.
While this year-long war comprises an extensive range of events, the narrator chose to present the WMD misinformation of Colin Powell as the precursor of the invasion.
One can argue why this is the case, but we argue that different narratives on the same event yield insights into consensual knowledge regarding the event.

In this paper we use and enhance \cite{kroll2020narratives} as a basis for our narrative definition.

\begin{definition}[Narratives]
    \label{def:narrative}
    A \emph{narrative} is defined as follows:
\begin{enumerate}
    \item Let $L$ be a set of literals. We call a directed edge-labeled graph (V, E) a \emph{narrative} with $V \subseteq \mathcal{E} \cup \Gamma \cup L$ being the nodes and $E \subseteq \mathcal{R}_F \cup \mathcal{R}_N$ being the edges.
    $\mathcal{R}_F$ is the set of \emph{factual relationships} that connect events and entities with entities and literals. $\mathcal{R}_N$ is the set of \emph{narrative relationships} that connect events with events.
    We denote the set of narratives as $\mathcal{N}$.
    \item For all $n \in \mathcal{N}$ we define a function $\eta(e, n)$ with $\eta: \Gamma \times \mathcal{N} \rightarrow \mathcal{N}$ that returns the narrative for event $e$ in $n$.
\end{enumerate}
\end{definition}

For our purpose, we can model narratives as directed graphs.
The nodes of the graphs are entities denoted by $\mathcal{E}$, event labels represented by $\Gamma$, and literals denoted by $L$.
Hence, narratives are able to connect events and their participants and enrich them with additional properties (expressed by labels)
We define two types of relationships.
Factual relationships, denoted by $\mathcal{R}_F$, are used to connect events and entities to entities or literals. 
They are used to describe relationships and roles of participants to events, and to enrich them by using literals.
In addition, we define relationships between events and events as narrative relationships, denoted by $\mathcal{R}_N$. 
Such relationships describe the \emph{unfolding of events} in the narrative, e.g., temporal and causal relationships.

\begin{figure}
    \centering
    \includegraphics[width=.45\textwidth]{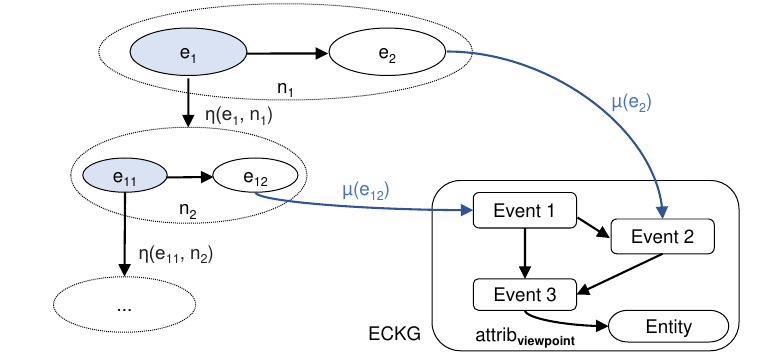}
    \caption{Narratives and their bindings to ECKGs.}
    \label{fig:param_pred_eckg}
\end{figure}

Each event $e \in \Gamma$ in a narrative $n \in \mathcal{N}$ is represented by a unique \emph{event label} with a concise description of the event.
Note that this label does not necessarily correspond to an event in an ECKG.
Referring to Fig.~\ref{fig:ukraine_invasion_example}, we can see this for \enquote{Belgrade Invasion}, an event that describes a part of the NATO bombing of Yugoslavia (represented in Wikidata by Q155723) but is not represented on its own. 
Additionally, each of those labels can be seen as a pointer to a narrative describing the event as denoted by $\eta(e, n)$. 
The $\eta$ function returns the narrative (i.e., a directed graph as described in (1)) that further describes $e$, as seen for \enquote{Redivision of the World} and \enquote{Iraq War} in our running example.
Additionally, this is illustrated in Fig.~\ref{fig:param_pred_eckg} with events $e_1$ and $e_{11}$.
We call nodes, for which $\eta$ yields a narrative a \emph{recursive node}.
For non-recursive nodes $ne$ we set $\eta(ne, n) = \emptyset$.

Furthermore, we introduce \emph{bindings} $\mu(e, n)$ for an event $e$ in narrative $n$.
While $\eta$ allows us to recursively traverse an event in the narrative by following the label to a sub-narrative, the binding $\mu$ maps the event label to its equivalent event in an ECKG.
This is illustrated in Fig.~\ref{fig:param_pred_eckg} with bindings of events $e_2$ and $e_{12}$ to a corresponding event in an ECKG.
Additionally, we can observe a binding in Fig.~\ref{fig:ukraine_invasion_example}, where $\mu(\text{\enquote{NATO Eastward Expansion}})$ provides us with the equivalent event in Wikidata (i.e., entity Q112127201).
That is, it returns the subgraph consisting of all triples with the identifier Q112127201 in the subject.
Note, that a label may in practice refer to several ids, i.e., homonyms might be a problem for $\mu$.
We will discuss difficulties in this regard in Sect.~\ref{subsec:narrative_bindings}.

As mentioned before, another key aspect of narratives is the narrator.
In this paper we simplify the narrator and represent her as a viewpoint.
Therefore, a narrative can differ depending on the viewpoint, e.g., regarding the events chosen to represent it.
With respect to Sect.~\ref{subsec:viewpoints}, this means that all information provided by a binding (here: all triples concerning the given event in an ECKG) must be viewpoint-compatible with the narrator.

%% file: sections/event_structures.tex
Based on the theoretical framework introduced in Sect.~\ref{sec:formalization} we can now introduce concrete event relationships that may be part of narratives.
We describe how narratives can be bound to ECKGs, i.e., how correspondences between events in narratives and events in ECKG can be defined. 
Finally, we introduce a method to construct narratives as synthesis from multiple text sources as one possible way of narrative mining.

\begin{figure*}
    \includegraphics[width=0.85\linewidth]{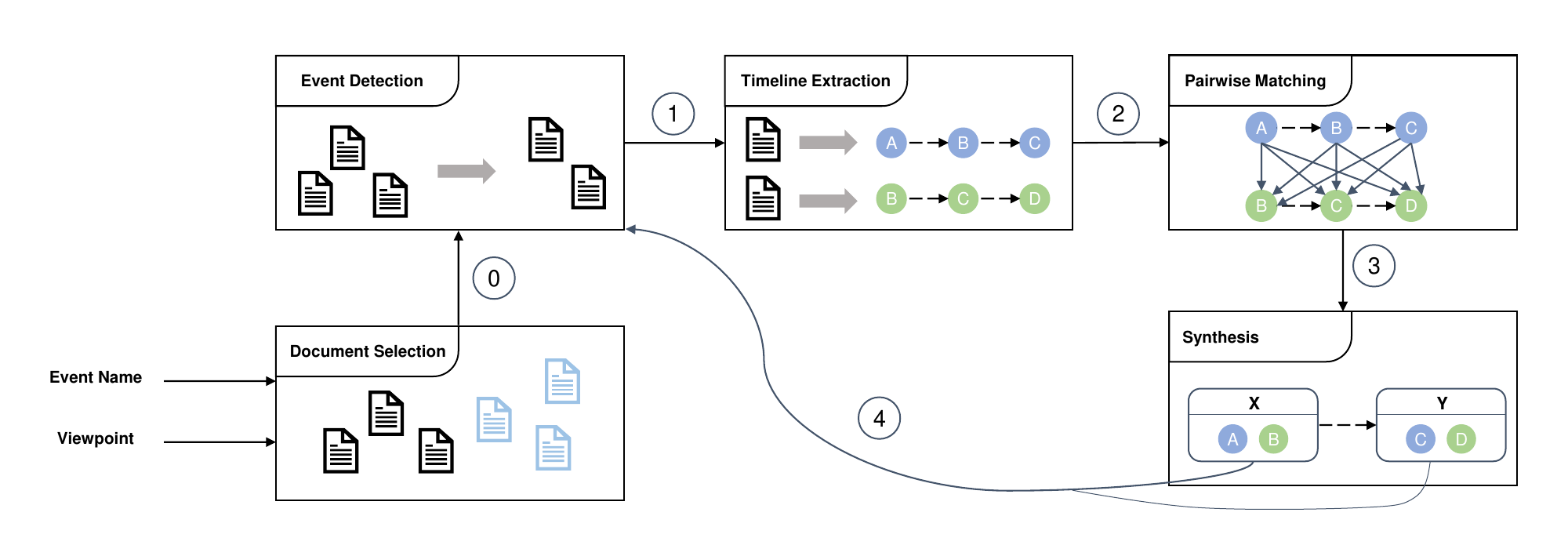}
    \caption{Sketch of our narrative mining algorithm.}
    \label{fig:rec_sum_algorithm}
\end{figure*}

\subsection{Narrative Relationships}
\label{subsec:event_relationships}
A key component in narratives are the narrative relationships $\mathcal{R}_N$.
We can further characterize such relationships as triples $\langle ev_1, p, ev_2 \rangle$ with $ev_1, ev_2 \in \Gamma$ and $p \in \Sigma_{NR}$ with $\Sigma_{NR}$ being a relation vocabulary.
Of course, the specific manifestation of $\Sigma_{NR}$ depends on the domain, i.e., the predicate labels used in narrative relationships must be defined by a domain expert.
However, we will define characteristics of a typical narrative relationship that can be used as guidance in this process.

We ground our deliberations in related work concerning event-event relationships \cite{hong2016eventeventrelations}.
Please note that we separate the vocabulary for constituent relationships and narrative relationships.
Although two events can form a narrative according to our definition, it is not sensible to interpret events in constituent relations as a narrative (otherwise each subevent relationship could be interpreted as narrative).
Hence, progress between both events should be described in order to define two events as a narrative. 
In general, we define three categories of narrative relationships:

\begin{description}
    \item[Temporal Relationships:] This category of relationships is concerned with the temporal order of events in a narrative. 
    Such relationships can be expressed by predicates like \emph{before}, \emph{after}, and \emph{during}.
    \item[Contingency and Causality:] The second category concerns causality and conditions for events.
    That is, this category comprises predicates like \emph{caused by}, \emph{lead to}, and \emph{has effect}.
    \item[Association:] The last category concerns associations, i.e., two events are associated on a general level.
    The corresponding labels are \emph{association} or \emph{associated with}.
\end{description}

According to Def.~\ref{def:events}, all events are associated with a time they took place, and hence, all events are in a temporal relationship with each other.
However, from a narrative point of view, some events are mentioned together, but their temporal relationship does not matter in the narrative.
Take, for instance, the association of the Belgrade Invasion, the Iraq War, and the War in Libya in Fig.~\ref{fig:ukraine_invasion_example}.
One could add temporal relationships between those three in the order mentioned before, but the temporal order does not matter in Putin's speech.
It is only necessary to connect those three events in the same context, i.e., the same recursive node. 

Additionally, those categories form a hierarchy, i.e., if there is a causal relationship between two events, they are also in a specific temporal relationship and are associated.
This is not true the other way around.
As mentioned before, while all events are, by definition, in a temporal relationship, the choice of relationship in the narrative depends on the narrator.
That is, associations are not superfluous as a category but can be used in cases where temporal and causal relationships are not known or not important.

\subsection{Binding Events in Narratives to ECKGs}
\label{subsec:narrative_bindings}
As discussed before, narratives can be grounded in ECKGs via a binding function $\mu$.
A bound narrative constructs a path in an ECKG by linking its event labels to knowledge graph events and connecting them by narrative relationships.
Hence, a binding can be interpreted as substituting the event label in the narrative by the respective subgraph in an ECKG.
If no corresponding event exists for an event label, the node in the narrative can be seen as virtual.

Binding the events mentioned in a narrative to named events in an ECKG provides two advantages.
First, it allows for disambiguation of the event, i.e., it is clear to which real-world event the narrative refers.
This eases the comparability between narratives since we compare knowledge graph entities instead of labels.
Second, the binding contextualizes the event by embedding it into a subgraph provided by the ECKG.
Furthermore, it includes additional metadata that can be used in downstream tasks like narrative comparison but may also include attributions from the ECKG  as long as they are viewpoint-compatible with the narrator.

The whole process can be seen as a variant of entity-linking or the recently introduced event linking \cite{yu2023eventlinking}. 
According to Def.~\ref{def:narrative} the nodes of a narrative are either events or entities.
If a node is an entity it must be linked to an event in a role relationship $r \in \mathcal{R}_R$.
In this case: if the corresponding event can be bound to an ECKG it is likely, that $r$ is also reflected in the ECKG.
In the case of event labels, we have to differentiate whether it is a recursive node or not.
We discuss the implications of both cases for the implementation of $\mu$ accordingly.

\emph{Non-recursive node}. As mentioned in Sect.~\ref{subsec:narratives}, events in a narrative are represented by labels like \enquote{Iraq War}.
Specifically, we talk about event labels $e$ with $\eta(e, n) = \emptyset$ here.
We can identify three cases, a binding function needs to account for:
\begin{description}
    \item[Direct Binding:] The event label can be directly and unambiguously matched to an event in an ECKG.
    In this case, the label can be substituted by the event identifier of the KG and treated as synonymous with the subgraph of the event (i.e., the set of all triples with this specific event as a subject in an ECKG). 
    For our running example in Fig.~\ref{fig:ukraine_invasion_example}, this is the case for many events, e.g., the Iraq War and the NATO eastward expansion.     
    \item[Indirect Binding:] In some cases, a node in a narrative can not be directly matched on an event in an ECKG but a similar one.
    This is the case for both the Belgrade Invasion and the collapse of the Soviet Union in our running example.
    \item[No Binding:] Suppose a node in a narrative represents not a named event, i.e., no corresponding event in an ECKG can be found. In that case, we substitute the label with a pointer to a subgraph that combines all known information of this event.
    In contrast to Direct Binding, this subgraph only exists in the narrative and is not materialized in the ECKG.
    For practical implementations, it might also be sensible to infer properties like the event type and a label for it.
\end{description}
The indirect binding case can become complicated since it requires a binding function to capture at least a part of the narrator's intention, i.e., the label should provide enough information in that regard.
For instance, in the Belgrade invasion, the mapped Wikidata entity represents the NATO bombing of Yugoslavia in 1999, which is close to what Putin wanted to express in his narrative.
The collapse of the Soviet Union, however, is linked to the entity representing the formal dissolution of the Soviet Union, which does not explicitly match the narrator's intention.
Deciding on a fitting entity in this specific case can hence be seen as a \emph{loss of focus} problem, i.e., the decision is whether a more general entity still encodes the intended semantic or is too general.

\emph{Recursive Node}. We evaluate $\eta(e, n)$ for recursive nodes and apply $\mu$ for each event in the generated narrative.
Afterward, we again apply $\eta$ for each event and continue recursively.
The recursion terminates if $\eta(e, n') = \emptyset$ for all $e$ in all narratives $n'$ traversed in the process. 

\subsection{Towards Multi-Viewpoint Narrative Mining}
\label{subsec:narrative_mining}
As discussed in Sect.~\ref{subsec:narratives}, our narratives resemble hierarchical event schemas.
In a recent work \cite{li2023schemainductionprompting} an incremental prompting approach was introduced to generate such schemas from a small document collection that contains instances of the respective event.
This particular approach has two advantages: First, it requires only slight supervision for the schema generation (i.e., only a small document collection), and second, it is domain-independent.

In this section, we introduce an algorithm based on \cite{li2023schemainductionprompting} for mining recursive narratives for a specific named event from a document collection.
The basic idea is to integrate the narrator in this setting, i.e., we assume a collection of documents for a given event $ev$ and a viewpoint $v$ to differ in terms of their narratives concerning $ev$.
That is, we do not try to abstract from specific scenarios to find event schema graphs but to find the main events $ev$ comprises according to $v$.
This step is performed recursively to generate narratives as depicted in Fig.~\ref{fig:ukraine_invasion_example}.
The algorithm is depicted in Fig.~\ref{fig:rec_sum_algorithm}.
A reference implementation and all prompts and data used in this paper are published on Github.\footnote{\url{https://github.com/fploetzky/WebSci2024}}
We discuss the algorithm step-by-step in the following.

\emph{Input and Output}. The algorithm takes an event $ev$ (e.g., the Iraq War or the Russo-Ukrainian War), a viewpoint $v$, and a document collection $D$ as an input.
The output of the algorithm is a narrative concerning $ev$ from the perspective of $v$ as reported in a subset $D_v \subseteq D$, i.e., all documents that can be seen as representative of $v$.
The algorithm utilizes a large language model (LLM) through prompting

\emph{Document Selection}. As a precursor, we determine the set $D_v$ in this step.
We follow \cite{ploetzky2022narrativeaspects} and proxy $v$ by a set of documents that describe $ev$.
Depending on $v$, $D_v$ could, for instance, be a set of newspaper articles from media outlets typically associated with $v$ or speeches by a specific speaker analogous to our running example.
This step can be done, e.g., by pre-selection and annotation of documents with the respective $v$ or by semi-automatically determining viewpoint clusters as illustrated in \cite{quraishi2018viewpoints}.

\emph{Event Detection}. After $D_v$ is determined, the actual first step of the algorithm is to detect, whether a document $d \in D_v$ actually refers to $ev$.
We prompt the LLM to verify whether an article is clearly about $ev$ and also provide a timespan $t$ for $ev$ in the prompt.
The timespan is used for disambiguation purposes for opaque event labels (for instance it is easy to mix the Gulf War in 1991 and the Iraq War in 2003 in an article).

\emph{Timeline Extraction}. The next step is analogous to the main event extraction in \cite{li2023schemainductionprompting}.
We generate a document event timeline $\tau(d)$ for each $d \in D_v$. An event timeline consists of all events in $d$ that are considered the main events concerning $ev$.
This is done in three steps:
(1) We use the prompt skeleton \emph{\enquote{Here is a news article: '\emph{\{text of $d$\}}' List the major events of the \emph{\{$ev$\}} in chronological order as reported in the article. Keep it consise and remove everything unrelated.}} to receive a list of events concerning $ev$ in $d$.
    (2) We use few-shot prompting to generate a precise description for each event in the list (i.e., a headline).
    This label should include a description and a point in time or time span describing when the event in question happened.
    More concrete, we utilized few-shot examples \enquote{\{\emph{event}\} -- \{\emph{event label} (time)\}} where \emph{event} is a sentence describing the event and \emph{event label} is a headline for this event including when it happened.
    (3) We again utilize few-shot prompting to 1. ensure that each event in the timeline is mentioned in $d$ (like in (1)) and 2. filter all events outside the timespan $t$ defined in Event Detection.

\emph{Pairwise Matching}. To generate a narrative, we have to unify all event timelines, i.e., for all $d \in D_v$ we combine the respective document timelines $\tau(d)$ and find connected components by clustering. 
For this, we conduct a pairwise comparison between the events of two event timelines $\tau(d_1)$ and $\tau(d_2)$ and repeat this process for each two documents in $D_v$.
We utilize SBERT~\cite{reimers2019sbert} to compute the Euclidean distance between each event label and HDBSCAN to cluster event labels in order to find the main events over all documents in $D_v$. 
Therefore, the result of this step is clusters of event labels describing the same event in the narrative.

\emph{Synthesis}.
The last step is to synthesize the event clusters from the previous step into a chain of events.
For each cluster, again, we generate a concise description, i.e., an event label on the basis of the event labels in the cluster via prompting.
For instance, in Fig.~\ref{fig:rec_sum_algorithm} in the Synthesis step, events A and B have been identified to be in a cluster during the pairwise matching and are now synthesized into event label X.
We additionally merge two clusters and prompt for a label if the labels have a cosine similarity below a given threshold.
Finally, we prompt for a label to determine the narrative relationships between all events in the timeline.

\emph{Recursion}. After constructing the narrative, we repeat the generation process for each node in it. 
In our example, we generate narratives for events A, B, C, and D, e.g., compute $\eta(A, X)$.
The recursion stops if $\eta(ev, n) = \emptyset$ for all events in all narratives.
To not lose the relationship between the original event and $ev$, we adapt the prompt in event detect to also include $ev$.

%% file: sections/proof_of_concept.tex
In this section, we conduct a detailed case study concerning the narrative model as introduced in Sect.~\ref{subsec:narratives} and its applications, the mining algorithm introduced in Sect.~\ref{subsec:narrative_mining}, and finally, event linking to Wikidata.

\subsection{Case Study and Document Collection}
\label{subsec:poc_use_case}
The case study concerns the Iraq War (2003-2011) as an event.
We chose this event as an example because 1. it generated international tension, and hence multiple viewpoints on this event exist, and 2. it is already completed and well-documented (in contrast to our running example, the Russo-Ukrainian War).

As viewpoints, we chose Russia, the United Kingdom, and the USA, i.e., $\mathcal{V} = \{\text{RU},\text{UK},\text{US}\}$. 
Both UK and US were directly involved in the war as allies.
In particular, the US were the driving force behind starting the war in the first place.
Russia was chosen because its government was skeptical of US claims and is naturally more reserved towards the United States.
Hence, those three viewpoints will most likely illustrate different narratives on the Iraq War.

\begin{figure*}
  \includegraphics[width=0.90\linewidth]{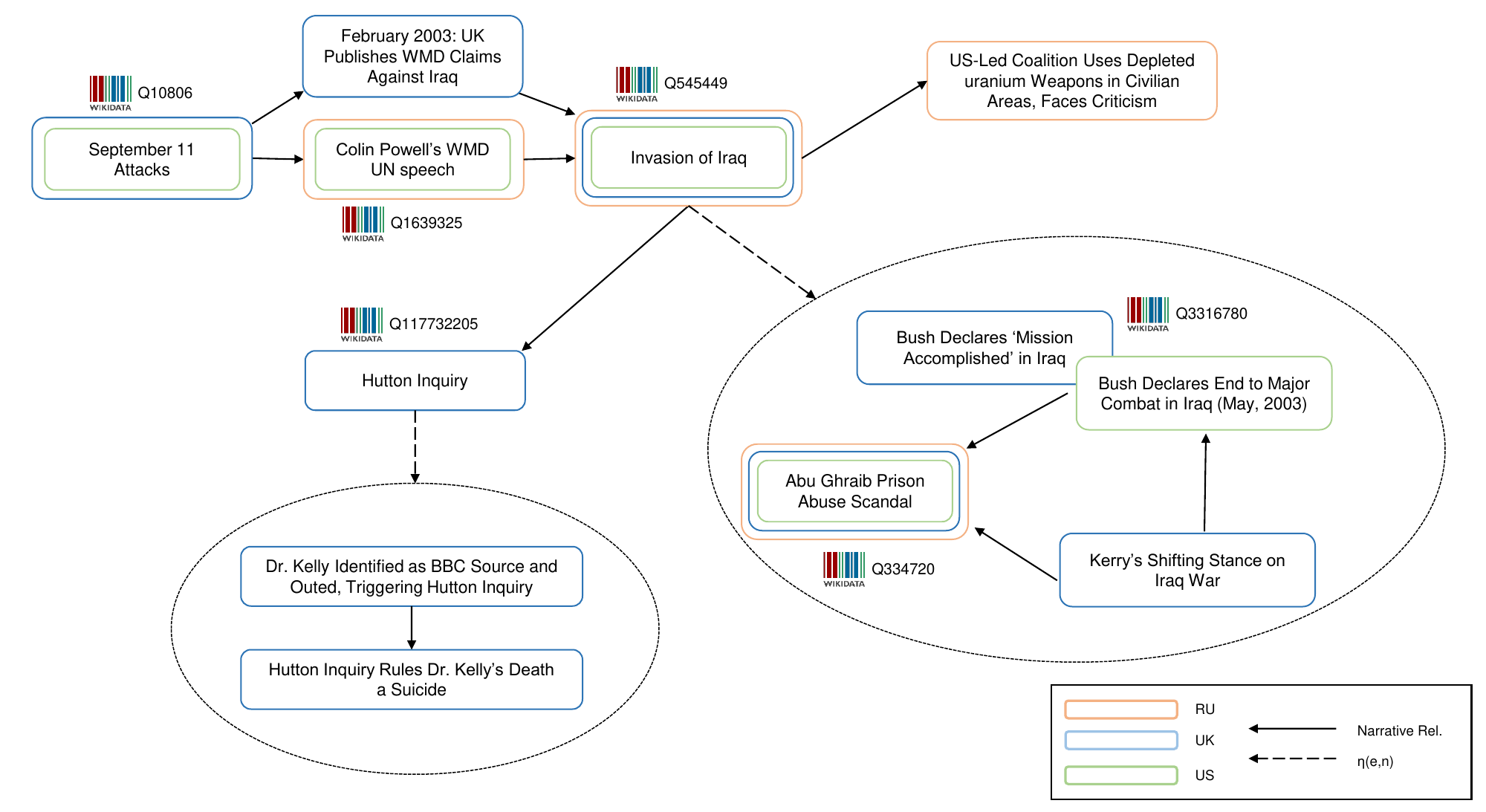}
  \caption{An excerpt of a narrative mined by our algorithm.}
  \label{fig:poc_narrative}
\end{figure*}

We designed the following research questions for our case study:

\begin{description}
  \item[RQ1] How is the quality of mined narratives from open-domain web data concerning the event labels and recursion?
  \item[RQ2] What are the potentials of narrative bindings, i.e., to which degree can we bind events in the mined narratives to Wikidata?
  \item[RQ3] What are the commonalities and differences between narratives with respect to the three viewpoints? 
\end{description}

Since this work introduces a novel view on events, no prior datasets exist that fit our needs to the best of our knowledge.
Hence, we collected a small dataset of newspaper articles concerning the Iraq War from well-known news outlets for each viewpoint.
This dataset is used as a proxy for the respective viewpoints.
Tab.~\ref{tab:stats} depicts an overview regarding the outlets chosen and how many articles were considered.
That is, for instance, for viewpoint RU we collected 197 documents from two news outlets.
We chose the outlets based on their popularity scores in Media Cloud collections \cite{roberts2021mediacloud} and whether an English version of the articles is available.
Otherwise, the outlet choice was random since the mining of narratives should be an abstraction of the underlying sources regardless of their structure.

Unfortunately, searching in Media Cloud for fitting articles was problematic since both the search interface and API have been under maintenance during the writing of this paper.
Instead, we used the Boolean query \enquote{iraq war 2003} on DuckDuckGo to find article candidates on the respective newspaper homepages.
For each viewpoint we considered the first seven DuckDuckGo result pages.
Additionally, we manually discarded articles whose URLs contained obvious references to other events as false positives (this was often the case in Russian newspapers that draw inferences from the Iraq War to the ongoing Russo-Ukrainian War).
Of course, this selection process could be more optimal and smaller in scale.
However, it shows the difficulties of event extraction and mining in real-world Web scenarios like heterogeneity of data, documents that only partially concern the desired topic, and no shared vocabulary.

\subsection{Mining and Binding Narratives}
\label{subsec:poc_mining_binding}

For this case study, we applied our mining algorithm to the document collections of all viewpoints.
All documents were pre-processed with newspaper3k\footnote{\url{https://newspaper.readthedocs.io/en/latest/}} to extract the title and body text.
To prevent articles that were too long or too short, we only considered articles between 1000 and 8000 characters.
We set the timespan to include all events between 2000 and 2012 so as not to miss potential war starters and effects that happened after the war.
After those two steps we ended with a total of 154 documents for RU, 188 documents for UK, and 207 documents for US. 
As a language model we utilized \llmUsed{}\footnote{\url{https://huggingface.co/meta-llama/Llama-2-70b-chat-hf}} for this study since it is a freely available state-of-the art LLM that runs on local hardware (in contrast to GPT3 used in \cite{li2023schemainductionprompting}).
Regarding SBERT we used all-mpnet-base-v2 as model.
HDBSCAN was utilized with an $\epsilon$ of $0.1$, a minimum cluster size of $10$, and a minimum of $2$ samples after experimentation with different parameter sets.
Due to the rather small number of documents, we showcase the narrative and the first recursion level. 
All experiments have been conducted on a server with 8 NVIDIA A40 GPUs, 2TB RAM, and an Intel(R) Xeon(R) Gold 6336Y CPU with 96 cores.
As an illustration, a portion of those three narratives, along with their bindings, is depicted in Fig.~\ref{fig:poc_narrative} and will be the basis of further discussion.
The colors surrounding the event labels indicate whether this event label was found in the respective viewpoints.
For readability, we unified the label if they occurred in multiple viewpoints.
We discuss \textbf{RQ1} with respect to the event labels, recursion, and narrative relationships separately.
Afterward, we evaluate \textbf{RQ2} accordingly.

\begin{table}
  \caption{Document count statistics}
  \label{tab:stats}
  \begin{tabular}{llr}
    \toprule
    \emph{Viewpoint} & \emph{Media Outlet} & \emph{\# Documents} \\
    \midrule
    \multirow{2}*{RU} &
    sputnikglobe.com & 59 \\
    &rt.com &  138 \\
    \midrule
    \multirow{2}*{UK} 
    & theguardian.com &  158 \\
    & independent.co.uk & 128 \\
    \midrule
    \multirow{3}*{US} 
    & latimes.com & 145 \\
    & washingtonpost.com & 59 \\
    & nytimes.com & 158 \\
    \bottomrule
  \end{tabular}
\end{table}

\emph{Nodes and Event Labels}. After applying our algorithm to the dataset described above, we ended up with three narratives comprising 16 event labels for US, 10 event labels for UK, and 7 for RU.
This reflects the size of $D_v$ for the three viewpoints as depicted in Tab.~\ref{tab:stats}.
The first recursion level totaled in 28 events for US, 35 for UK, and 9 for RU; a major increase for UK due to one of the events in the UK narrative being the Iraq invasion. 
Therefore, in total our mining algorithm generated 105 event labels. 

A major obstacle for our mining algorithm is the concise description of events.
We explicitly decoupled mining and binding events to allow the algorithm to be as domain-independent as possible.
This also means that no shared vocabulary exists for the event labels, and hence, this can be seen as a variant of open domain event extraction \cite{xiang2019eventsurvey}.
However, besides this difficulty, the quality of the event labels turned out to be good when it comes to the actual label and worthy of improvement when it comes to the extracted time intervals.
Regarding the quality of the event labels, consider \enquote{Hutton Inquiry Rules Dr. Kelly's Death a Suicide} and \enquote{Abu Ghraib Prison Abuse Scandal} in Fig.~\ref{fig:poc_narrative} as representatives of all other labels.

However, as mentioned, when it comes to the time of an event, was is oftentimes not possible to extract an exact time, or the time was simply wrong.
An example of the latter is \enquote{May 2017: High Court Rules in Favor of Prosecution in 'R v Jones' Case, UK Military Leaders Consider War Crimes Immunity for Iraq Invasion during the Iraq War} from the UK viewpoint.
While the long label is correct and the event covers the Iraq War, the court case took place in 2006 and not 2017 \cite{osborne2017highcourtuk}.
Concerning the connotation of the Iraq War, we only observed one event label out of all 105 events that did not belong to the Iraq War, namely the Russian annexation of Ukraine that was mentioned in the UK in one instance.

\emph{Narrative Relationships}. In this proof of concept, we limited ourselves to one temporal relationship, specifically \enquote{happened after}.
However, generating those labels proved to be the most inconsistent part of the algorithm.
First, the results of prompting the LLM to generate event timestamps were only partially successful and oftentimes erroneous, as mentioned before.
Future approaches may, therefore, utilize external tools for this task.
Second, even with correct time annotations in the event labels, drawing the correct \enquote{happened after} relationship turned out to be a challenging task in our algorithm for the LLM.

\emph{Recursion}.
The results of the recursion step depended heavily on the event label generated during the first run of the algorithm.
For instance, for the UK, one of the events mined for the Iraq War was the Invasion of Iraq as depicted in Fig.~\ref{fig:poc_narrative}.
For this event alone, 15 events have been created during the recursion step.
That is, a more general label will naturally produce narratives with more events.
In some cases, the general label is just refined.
For instance, the narrative for the Abu Ghraib prison abuse scandal in Fig.~\ref{fig:poc_narrative} includes an event \enquote{US Military Personnel Convicted for Abu Ghraib Prison Abuse, 2004}.

In summary, for \textbf{RQ1}, we conclude that the event abstraction and recursion parts in our mining algorithm can produce hierarchical narratives.
Bindings and the utility of such narratives are studied in the remaining research questions.
The main obstacle here is the extraction of event times and the narrative relationships.
For this study, we limited ourselves to only a single relationship, but even this rather simple temporal relationship proved to be problematic.
Improvements should hence focus on this part first.

\emph{Event Linking to Wikidata}.
For \textbf{RQ2}, we studied the binding of events in our narratives to Wikidata as our ECKG choice.
We have chosen Wikidata due to its scale and popularity as the largest freely available knowledge graph.
We did not implement a binding algorithm for this proof of concept.
Instead, we focused on detecting potential problems such an algorithm would face with respect to Sect.~\ref{subsec:narrative_bindings} and recent event linking literature \cite{ou2023hierarchicaleventgrounding,yu2023eventlinking}.
For the bindings, we used Wikipedia as a proxy to search for titles similar to the event label we tried to bind.
That is, we used the event label generated by our algorithm and searched for a fitting Wikipedia article title or chapter title in an article with a title closely related to the event label.
The latter was done to cater to the aforementioned indirect binding case, e.g., to find the Belgrade Invasion in the Wikipedia article concerning the NATO bombing of Yugoslavia in 1999.
We chose Wikipedia as a proxy because 1. recent works in event linking identified Wikipedia as a suitable linking destination \cite{yu2023eventlinking}, and 2. Wikipedia articles often have a corresponding entity in Wikidata that shares the same title.
Fig.~\ref{fig:poc_narrative} annotates the Wikidata identifier for all events for which binding was possible.
In summary, we have been able to bind 22 of the 44 events in US, 22 out of 45 for UK, and 6 out of 16 for RU (i.e., 50 out of 105 in total) to Wikidata.
Note that this can not be seen as a binding quality indicator since, naturally, not all events in a narrative will have a counterpart in an ECKG, as discussed before.
We identified, however, several problems, in particular:
(1) compound event labels, i.e., event labels that are actually referring to two events and should hence be split. 
(2) Double matches in bindings, for instance, \enquote{US and Iraqi Forces Launch Offensives in Fallujah, 2004} which can refer to two events in Wikidata since there have been two battles in Fallujah in 2004 (i.e., the timestamp 2004 is too general). 
(3) The indirect binding case as discussed in Sect.~\ref{subsec:narrative_bindings} was observed multiple times.
For instance, for \enquote{UN Issues Iraq Ultimatum: Disarm or Face Military Action (2002)} where a possible binding could be the whole relationship of the UN security council regarding the Iraq War (Q7888917). 
However, this does not only include the Iraq Ultimatum but also different events like the Blix investigations.
The opposite case can be observed for the recursive node \enquote{Invasion of Iraq} in Fig.~\ref{fig:poc_narrative}, which should refer to the War in Iraq (Q545449) and not only the actual invasion (Q107802) when considering the narrative generated from this node (the Abu Ghraib prison abuse scandal happened after the invasion but in the Iraq War).
Hence, binding a recursive node requires an understanding of its inner structures.
Regarding our previous assumption of connecting events that are otherwise unconnected in ECKGs, this turns out to be true for some events.
For instance, Colin Powell's address is represented in Wikidata (Q1639325) but is not connected to the Iraq War.
Even if it was connected via a part of the relationship, the prominent role this event played for some viewpoints is only reflected in the narrative.

We conclude the discussion of \textbf{RQ2} and remark that binding a recursive narrative to an ECKG is a non-trivial task that requires in-depth analysis in future work.

\subsection{Narrative Comparison}
\label{subsec:narrative_comparison}
In this section, we discuss \textbf{RQ3} and compare the different narratives generated for this case study.
Additionally, we hint at future applications for our narrative model.

\emph{Commonalities}. As hypothesized in the introduction, different narratives of an event will highlight different aspects of it.
Before focusing on the difference, we first investigated which events are shared by all three narratives.
Examples include the actual invasion of Iraq and the Abu Ghraib prison scandal.
However, the latter was featured in the main narrative in US and RU, but only after the first recursion in UK (in the invasion of Iraq event; hence, it is located there in Fig.~\ref{fig:poc_narrative}).
In some cases, the narratives shared the same event but with different labels, e.g., George W. Bush's \enquote{Mission Accomplished} speech in US and UK (overlapping in Fig.~\ref{fig:poc_narrative}).
Also, all narratives mention at some point the WMDs that were allegedly possessed by the Iraqi government.

\emph{Differences}. We start the differences with exactly those WMDs. 
While in RU and US they are introduced in Colin Powell's address to the UN security council (analogous to Putin's argumentation in Fig.~\ref{fig:ukraine_invasion_example}), UK media highlights the WMD claim made in a document from the British government (wrongly dated by the algorithm).
Overall, the UK viewpoint focused more often on British entities, especially former British Prime Minister Tony Blair.
Note that this can be a sample bias due to the small document collection and only two outlets, but it was a noteworthy observation.
Hence, the model identifies the core actors and their role in the narrative.

Another main difference is the start of the narrative, i.e., the initial reason for this war.
For US and UK, the 9/11 attacks have mostly been mentioned as the start of the war, while RU mentions the Colin Powell speech as its start.
Of course, this can not be generalized in this small proof of concept, but it at least indicates a framing (i.e., reaction to a terror attack versus a false assumption as a war start).
Following this observation allows us to derive different paths a story took from a specific point of view.

Finally, and most notably, the narratives generated during the recursion steps showed the potential to include specific events and frames that are unique to a specific narrator.
We highlight two examples.
First, Fig.~\ref{fig:poc_narrative} contains a narrative concerning the Hutton Inquiry for UK.
This inquiry was conducted regarding the death of Dr. David Kelly, who provided anonymous information to the BBC regarding the overstatement of the dangers of WMDs as published in the government document mentioned above.
Since this case is not well-known outside of UK it makes sense that it is not included in the other two viewpoints.
Second, RU features an event of the US-led coalition using depleted uranium ammunition in Iraq (visible in Fig.~\ref{fig:poc_narrative}).
The narrative behind this event actually connects the usage of depleted uranium ammunition with an increase in congenital birth defects in Fallujah (after the aforementioned invasion).
After researching our corpus, the connection between the usage of depleted uranium and birth defects is actually available in Russian media and was correctly identified and categorized by our algorithm.
Capturing this small narratives and connecting them to a greater one, i.e., categorizing them, can be seen as the major future direction for utilizing narratives for complex events.

\subsection{Limitations}
\label{subsec:poc_discussion}
In this section, we briefly discuss some significant limitations of our approach that can not be evaluated in this proof of concept.

\emph{Hallucinations and Stochasticity}. Hallucinations are always a problem when dealing with LLMs (cf. \cite{pride2023coregpt}).
We did not observe any hallucinations in our scenario beyond wrong dates for events (i.e., all events have been mentioned in at least one document of the respective viewpoint), but this observation can not be generalized.
Additionally, due to the stochastic nature of LLMs, narrative mining is not guaranteed to yield the same graphs.
In our small-scale tests, only minor changes in event labels occurred, but on a larger scale, this must be evaluated.

\emph{Scalability}. In this paper we focussed on adopting \cite{li2023schemainductionprompting} for our narrative mining algorithm.
The main advantage here is the generation of narratives without any setup or preprocessing except by providing the documents.
However, this method does not scale in general.
The generation of the US graph and the first recursion step takes about 5 hours on a high-end server and uses parallelization and quantization to perform the prompting.
Hence, the algorithm must be considered preliminary work but needs additions in future work for actual productive usage.

%% file: sections/related_work.tex
\emph{Computational Narratives}. Research on utilizing narratives as a method for information representation and understanding was conducted in different variations over the last years (cf.~\cite{ranade2022surveycomputationalunderstandingofnarratives} for a survey).
However, what a narrative actually \emph{is} varies in-between literature \cite{norambuena2023surveyeventnarratives}.
Hence, various models of narratives have been discussed, focussing on different narrative aspects.
Connecting-the-dots techniques for constructing storylines out of news headlines \cite{shahaf2011connectingthedotsnews,norambuena2020narrativemaps,yan2023narrativegraphevolving} are similar to this work in their notion of event chains.
In contrast to our work, however, hierarchical aspects, i.e., different levels of granularity, are not discussed.
Additionally, our work is focused on whole documents and not only headlines.
Utilizing multiple viewpoints for narratives have also been modeled in previous works \cite{lakoff2010computationalnarratives} but to best of our knowledge never used to compare different narratives and link them to knowledge graphs.
The last category of related work concerns applications that utilize narratives, for instance, for news retrieval \cite{voskarides2021newsretrievaleventnarratives} or social media analysis \cite{dash2022indiantwitternarratives}.
In contrast to this paper, both categories focus on concrete tasks and use narrow definitions of narratives tailored for those.

\emph{Event Models and Schemas}. Modeling events have been a topic for at least 40 years \cite{chen2020explainableevents}. They have been modeled as simple schemas (e.g. ACE), cubes \cite{li2017eventcube}, or by utilizing first order logic \cite{guizzardi2013towardsontologicalevents,almeida2019eventsasentities}.
Additionally, event models like SEM \cite{vanhage2011sem}, the F-Model \cite{scherp2009eventmodelf}, LODE~\cite{shaw2009lodemodel}, and FARO~\cite{rebboud2022eventrelationsinkgs}  have been proposed to model various aspects of events either for specific domains or in general (cf. \cite{piryani2023eventontologysurvey} for a recent comparison of such event ontologies).
Another related topic in this regard is event schema induction, i.e., the task of learning a typical schema for an event label in an unsupervised fashion. 
In the past, different approaches have been investigated either to learn event role-slot templates \cite{chambers2011templatelearning} or chains of events \cite{chambers2008narratives,chambers2009narrativeschemas}.
Recently, the notion of event schema graphs has been proposed \cite{li2020eventschemagraphs} that add the idea of connecting relationships between events and participants of events.
Induction methods for such graphs have been proposed by utilizing specialized neural architectures \cite{li2020eventschemagraphs,li2021complexeventschemainduction} or by prompting techniques for large language models such as GPT3 \cite{li2023schemainductionprompting}.

\emph{Event Linking and Grounding}. Since events are modeled as entities in ECKGs, techniques applied in entity linking tasks can also be used for event linking.
However, some works are specifically dedicated to linking events, for instance to Wikipedia~\cite{yu2023eventlinking} or Wikidata~\cite{ou2023hierarchicaleventgrounding}. 
The latter work specifically considers hierarchical event grounding, i.e., event linking to complex events in Wikidata on different levels of granularity.
\balance

%% file: sections/conclusion.tex
This paper introduced narratives as rich event semantics for complex real-world events.
This includes connecting events to represent the inner structures of more complex events and considering multiple points of view.
We argue that it becomes increasingly important to consider multiple viewpoints on real-world events in our current polarizing times.
This is especially true since we observe the world primarily through intermediaries' lenses, i.e., by different narratives from various narrators.
Therefore, in this paper, we unified the concepts of viewpoints, narratives, event-centric knowledge graphs into a conceptual model.

Connecting narratives mined from unstructured discourse to structured knowledge repositories is one way of achieving multiple viewpoints on events in knowledge graphs.
Based on the results of our proof of concept, we argue that the recent advances in natural language processing allow us to mine narratives from different sources in acceptable quality.
However, this paper can only be considered as a starting point.
Further research is necessary, especially concerning the practical usage of the proposed mining algorithm and binding events in narratives to ECKGs.